\documentclass[runningheads]{llncs}
\usepackage{amssymb} %
\usepackage{amsmath} %Math package - small matrices
\usepackage{bm}

\usepackage{graphicx} %For images
\newsavebox\CBox
\def\textBF#1{\sbox\CBox{#1}\resizebox{\wd\CBox}{\ht\CBox}{\textbf{#1}}}
\usepackage{epstopdf}

\usepackage{algpseudocode}
\usepackage{algorithm}

\usepackage{multirow}

\usepackage{hyperref}
\hypersetup{colorlinks   = true, %Colours links instead of ugly boxes
linkcolor    = red, %Colour of internal links
citecolor   = blue  %Colour of citations
}

\usepackage{booktabs}

\newif\ifalgo
\algofalse

\begin{document}

%\setlength{\abovedisplayskip}{4.5pt}
% \setlength{\belowdisplayskip}{4.5pt}

% \setlength\abovecaptionskip{2pt}
% \setlength{\abovecaptionskip}{15pt plus 3pt minus 2pt}

%%%%%%%%%%%%%%%%
% Title
%%%%%%%%%%%%%%%%
\title{DeepASL: Kinetic Model Incorporated Loss for Denoising Arterial Spin Labeled MRI via Deep Residual Learning}

\titlerunning{DeepASL: Kinetic Model Incorporated Loss for Denoising ASL}

\author{Cagdas~Ulas$^{1}$, Giles~Tetteh$^{1}$, Stephan~Kaczmarz$^{2}$, Christine Preibisch$^{2}$, and Bjoern~H.~Menze$^{1}$}
\authorrunning{Ulas et al.}

\institute{%
$^1$ Department of Computer Science, Technische Universit\"{a}t M\"{u}nchen, Germany\\
$^2$ Department of Neuroradiology, Technische Universit\"{a}t M\"{u}nchen, Germany}

\maketitle

%%%%%%%%%%%%%%%%
% Abstract
%%%%%%%%%%%%%%%%
\begin{abstract}
Arterial spin labeling (ASL) allows to quantify the cerebral blood flow (CBF) by magnetic labeling of the arterial blood water. ASL is increasingly used in clinical studies due to its noninvasiveness, repeatability and benefits in quantification. However, ASL suffers from an inherently low-signal-to-noise ratio (SNR) requiring repeated measurements of control/spin-labeled (C/L) pairs to achieve a reasonable image quality, which in return increases motion sensitivity. This leads to clinically prolonged scanning times increasing the risk of motion artifacts. Thus, there is an immense need of  advanced imaging and processing techniques in ASL. In this paper, we propose a novel deep learning based approach to improve the perfusion-weighted image quality obtained from a subset of all available pairwise C/L subtractions. Specifically, we train a deep fully convolutional network (FCN) to learn a mapping from noisy perfusion-weighted image and its subtraction (residual) from the clean image. Additionally, we incorporate the CBF estimation model in the loss function during training, which enables the network to produce high quality images while simultaneously enforcing the CBF estimates to be as close as reference CBF values. Extensive experiments on synthetic and clinical ASL datasets demonstrate the effectiveness of our method in terms of improved ASL image quality, accurate CBF parameter estimation and considerably small computation time during testing.
\end{abstract}

%%%%%%%%%%%%%%%%
% Main Text
%%%%%%%%%%%%%%%%

%%%%%%
%Introduction
\section{Introduction} \label{sec:Intro}
Arterial spin labeling (ASL) is a promising MRI technique that allows quantitative measurement of cerebral blood flow (CBF) in the brain and other body organs. ASL-based CBF shows a great promise as a biomarker for many neurological diseases such as stroke and dementia, where perfusion is impaired, and thereby the blood flow alterations need to be investigated \cite{Alsop}. ASL has been increasingly used in clinical studies since it is completely non-invasive and uses magnetically labeled blood water as an endogenous tracer where the tagging is done through inversion radio-frequency (RF) pulses \cite{Alsop,Spann}. In ASL, a perfusion-weighted image is obtained by subtracting a label image from a control image in which no inversion pulse is applied. The difference reflects the perfusion, which can be quantified via appropriate modelling \cite{Alsop,Owen}.

Despite its advantages, ASL significantly suffers from several limitations including the low signal-to-noise ratio (SNR), poor temporal resolution and volume coverage in conventional acquisitions \cite{Fang}. Among these limitations, the low SNR is the most critical one, necessitating numerous repetitions to achieve accurate perfusion measurements. However, this leads to impractical long scanning time especially in multiple inversion time (multi-TI) ASL acquisitions with increased susceptibility to motion artifacts \cite{Spann,Alsop,Kim}. 

To alleviate this limitation, several groups have proposed spatial and spatio-temporal denoising techniques, for instance denoising in the wavelet domain \cite{Bibic}, denoising in the image domain using adaptive filtering \cite{Wells}, non-local means filtering combined with wavelet filtering \cite{Liang}, spatio-temporal low-rank total variation \cite{Fang}, and spatio-temporal total generalized variation \cite{Spann}. Just recently, a deep learning based ASL denoising method \cite{Kim} has been shown to produce compelling results. All of these methods primarily consider improving the quality of noisy perfusion-weighted images, followed by CBF parameter estimation as a separate step although accurate quantification of CBF is the main objective in ASL imaging.

In this paper, unlike the previous deep learning work \cite{Kim} which is only data driven, we follow a mixed modeling approach in our denoising scheme. In particular, we demonstrate the benefit of incorporating a formal representation of the underlying process -- a CBF signal model -- as a prior knowledge in our deep learning model. We propose a novel deep learning based framework to improve the perfusion-weighted image quality obtained by using a lower number of subtracted control/label pairs. First, as our main contribution, we design a custom loss function where we incorporate the Buxton kinetic model \cite{Buxton} for CBF estimation as a separate loss term, and utilize it when training our network. Second, we specifically train a deep fully-convolutional neural network (CNN) adopting the residual learning strategy \cite{He}. Third, we use the images from various noise levels to train a single CNN model. Therefore, the trained model can be utilized to denoise a test perfusion-weighted image without estimating its noise level. Finally, we demonstrate the superior performance of our method by validations using synthetic and clinical ASL datasets. Our proposed method may facilitate scan time reduction, making ASL more applicable in clinical scan protocols.

%%%%%%
%Methods
\section{Methods} 
\label{sec:Methods}

\subsection{Arterial Spin Labeling} 
\label{sec:ASL}
In ASL, arterial blood water is employed as an endogenous diffusible tracer by inverting the magnetization of inflowing arterial blood in the neck area by using RF pulses. After a delay for allowing the labeled blood to perfuse into the brain, label and control images are repeatedly acquired with and without tagging respectively \cite{Alsop,Owen}. The signal difference between control and label images is proportional to the underlying perfusion \cite{Alsop}. The difference images are known as perfusion-weighted images ($\mathrm{\Delta} \mathrm{M}$), and can be directly used to fit a kinetic model. For CBF quantification in a single inversion-time (TI) ASL, the single-compartment kinetic model (so-called Buxton model \cite{Buxton}) is generally used. According to this model, the CBF in $\mathrm{ml/100 g/min}$ can be calculated in every individual voxel for pseudo-continuous ASL (pCASL) acquisitions as follows,
\begin{equation}
f(\mathrm{\Delta} \mathrm{M}) = \mathrm{CBF} =\frac{6000 \cdot \beta \cdot \mathrm{\Delta} \mathrm{M} \cdot e^{\frac{PLD}{T_\mathrm{1b}}}}{2 \cdot \alpha \cdot T_\mathrm{1b}  \cdot \mathrm{SI}_\mathrm{PD} \cdot \left(1-e^{-\frac{\tau}{T_\mathrm{1b}}} \right)},
\label{BuxtonModel}
\end{equation}
where $\beta$ is the brain-blood partition coefficient, $T_{\mathrm{1b}}$ is the longitudinal relaxation time of blood, $\alpha$ is the labeling efficiency, $\tau$ is the label duration, $PLD$ is the post-label delay, and $\mathrm{SI}_{\mathrm{PD}}$ is the proton density weighted image \cite{Alsop}.
%\subsection{Problem Formulation and Loss Function Design}
%\label{sec:loss-function}

\subsection{Deep Residual Learning for ASL Denoising}
\label{sec:Residual-learning}

\subsubsection{Formulation.} 
\label{sec:formulation}
Our proposed CNN model adopts the residual learning formulation \cite{He,Kiku}. It is assumed that the task of learning a residual mapping is much easier and more efficient than original unreferenced mapping \cite{Zhang}. With the utilization of a residual learning strategy, extremely deep CNN can be trained and superior results have been achieved for object detection \cite{He} and image denoising \cite{Zhang} tasks.

The input of our CNN model is a noisy perfusion-weighted image $\mathrm{\Delta} \mathrm{M}_n$ that is obtained by averaging a small number of pairwise C/L subtractions. We denote a complete perfusion-weighted image as $\mathrm{\Delta} \mathrm{M}_c$ estimated by averaging all available C/L subtractions. We can relate the noisy and complete perfusion-weighted image as $\mathrm{\Delta} \mathrm{M}_n = \mathrm{\Delta} \mathrm{M}_ c + \mathrm{N}$, where $\mathrm{N}$ denotes the noise image which degrades the quality of the complete image. Following the residual learning strategy, our CNN model aims to learn a mapping between $\mathrm{\Delta} \mathrm{M}_n$ and $\mathrm{N}$ to produce an estimate of the residual image $\tilde{\mathrm{N}}$; $\tilde{\mathrm{N}} = \mathcal{R}(\mathrm{\Delta} \mathrm{M}_n| \mathbf{\Theta})$, where $\mathcal{R}$ corresponds to the forward mapping of the CNN parameterised by trained network weights $\mathbf{\Theta}$. The final estimate of the complete image is obtained by $\mathrm{\Delta} \tilde{\mathrm{M}}_c = \mathrm{\Delta} \mathrm{M}_n - \tilde{\mathrm{N}} $.

% \vspace{-4mm}
\subsubsection{Loss Function Design.} \label{sec:lossfunction}
In this work, we design a custom loss function to simultaneously control the quality of the denoised image and the fidelity of CBF estimates with respect to reference CBF values. Concretely, given a set of training samples $\mathcal{D}$ of input-target pairs ($\mathrm{\Delta} \mathrm{M}_n, \mathrm{N}$), a CNN model is trained to learn the residual mapping $\mathcal{R}$ for accurate estimation of complete image by minimizing the following cost function, 
\begin{equation}
 \mathcal{L}(\mathbf{\Theta}) = \sum_{(\mathrm{\Delta} \mathrm{M}_n, \mathrm{N}) \in \mathcal{D}}\lambda\|\mathrm{N}-\tilde{\mathrm{N}}\|_2^2 \enskip + \enskip (1-\lambda)\|\mathrm{f}_t - f(\mathrm{\Delta} \mathrm{M}_n -\tilde{\mathrm{N}}; \mathbf{\xi})\|_2^2,
 \label{lossfunction}
\end{equation}
where $\lambda$ is regularization parameter controlling the trade-off between the fidelity of the residual image and CBF parameter estimates, $\mathrm{f}_t$ is the reference CBF value for each voxel, and $\mathbf{\xi}$ denotes all the predetermined variables as given in (\ref{BuxtonModel}). We emphasize that the second term of our loss function (\ref{lossfunction}) explicitly enforces the consistency of CBF estimates with respect to reference CBF values, computed from the complete perfusion-weighted image through the use of the Buxton kinetic model. This integrates the image denoising and CBF parameter estimation steps into a single pipeline allowing the network to generate better estimates of perfusion-weighted images by reducing noise and artifacts.

% \vspace{-4mm}
\subsubsection{Network Architecture.}
\label{sec:networkarchitecture}

\begin{figure}[t!]
 \centering
 \includegraphics[width=0.95\columnwidth]{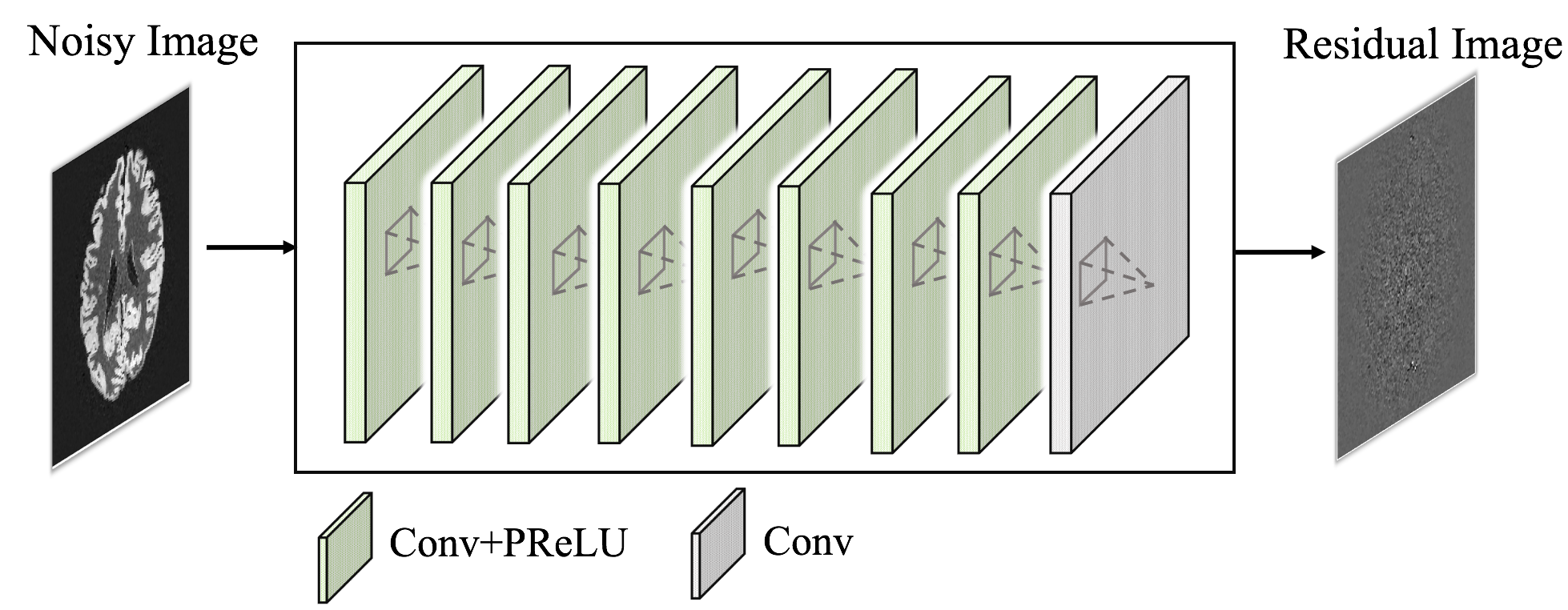}
 \caption{The architecture of the proposed network used for the estimation of a residual image from the noisy perfusion-weighted image given as input. \label{fig:NetworkArchitecture}}
\end{figure}

Fig.~\ref{fig:NetworkArchitecture} depicts the architecture of our network. The network takes 2D noisy gray image patches as input and residual image patches as output. Our network consists of eight consecutive 2D convolutional layers followed by parametric rectified linear units (PReLU) activation. Although ReLU activation has been reported to achieve good performance in denoising tasks \cite{Kim,Zhang}, we empirically obtained better results on our ASL dataset using PReLU in which negative activation is allowed through a small non-zero coefficient that can be adaptively learned during training \cite{He2015}. The number of filters in every convolutional layer is set to 48 with a filter size of $3 \times 3$. Following eight consecutive layers, we apply one last convolutional layer without any activation function. The last layer only includes one convolutional filter, and its output is considered as the estimated residual image patch.

% \vspace{-4mm}
\subsubsection{Training.}
\label{sec:training}
Training was performed using $18000$ noisy and residual patch pairs of size $40 \times 40$. The network was trained using the Adam optimizer with a learning rate of $10^{-4}$ for 200 epochs and mini-batch size of 500. We trained a single CNN model for denoising the noisy input images from different noise levels. % The noisy images were generated by averaging the subtractions of C/L from a subset of all available pairs. 

\section{Experiments and Results}
\subsubsection{Datasets.}
\label{sec:Dataset}
Pseudo-continuous ASL (pCASL) images were acquired from $5$ healthy subjects on a 3T MR scanner with a 2D EPI readout using the following acquisition parameters (TR/TE = $5000/14.6$ $\text{ms}$, flip angle = $90^{\circ}$, voxel size = $2.7 \times 2.7 \times 5$ $\text{mm}^3$, matrix size = $128 \times 128$, $17$ slices, labeling duration ($\tau$) = $1800$ $\text{ms}$, post-label delay ($PLD$) = $1600$ $\text{ms}$). $30$ C/L pairs and one $ \mathrm{SI}_\mathrm{PD}$ image were acquired for each subject.  

Additionally, high resolution synthetic ASL image datasets were generated for each real subject based on the acquired $ \mathrm{SI}_\mathrm{PD}$ and coregistered white-matter (WM) and grey-matter (GM) partial volume content maps. To create a ground-truth CBF map, we assigned the CBF values of $20$ and $65$ $\mathrm{mL/100g/min}$ to the WM and GM voxels respectively, as reported in \cite{Spann}. To generate synthetic data with a realistic noise level, the standard deviation over 30 repetitions was estimated from the acquired C/L images for each voxel. We subsequently added Gaussian noise with estimated standard deviation to each voxel of the synthetic images. This step was repeated 100 times to create a synthetic data per subject containing 100 C/L pairs. For synthetic data, we set $\tau = 1600$ $\text{ms}$ and $PLD = 2200$ $\text{ms}$. All the other constant variables in (\ref{BuxtonModel}) were fixed based on the recommended values for pCASL given in \cite{Alsop}.

\begin{figure}[t!]
 \centering
 \includegraphics[width=\columnwidth]{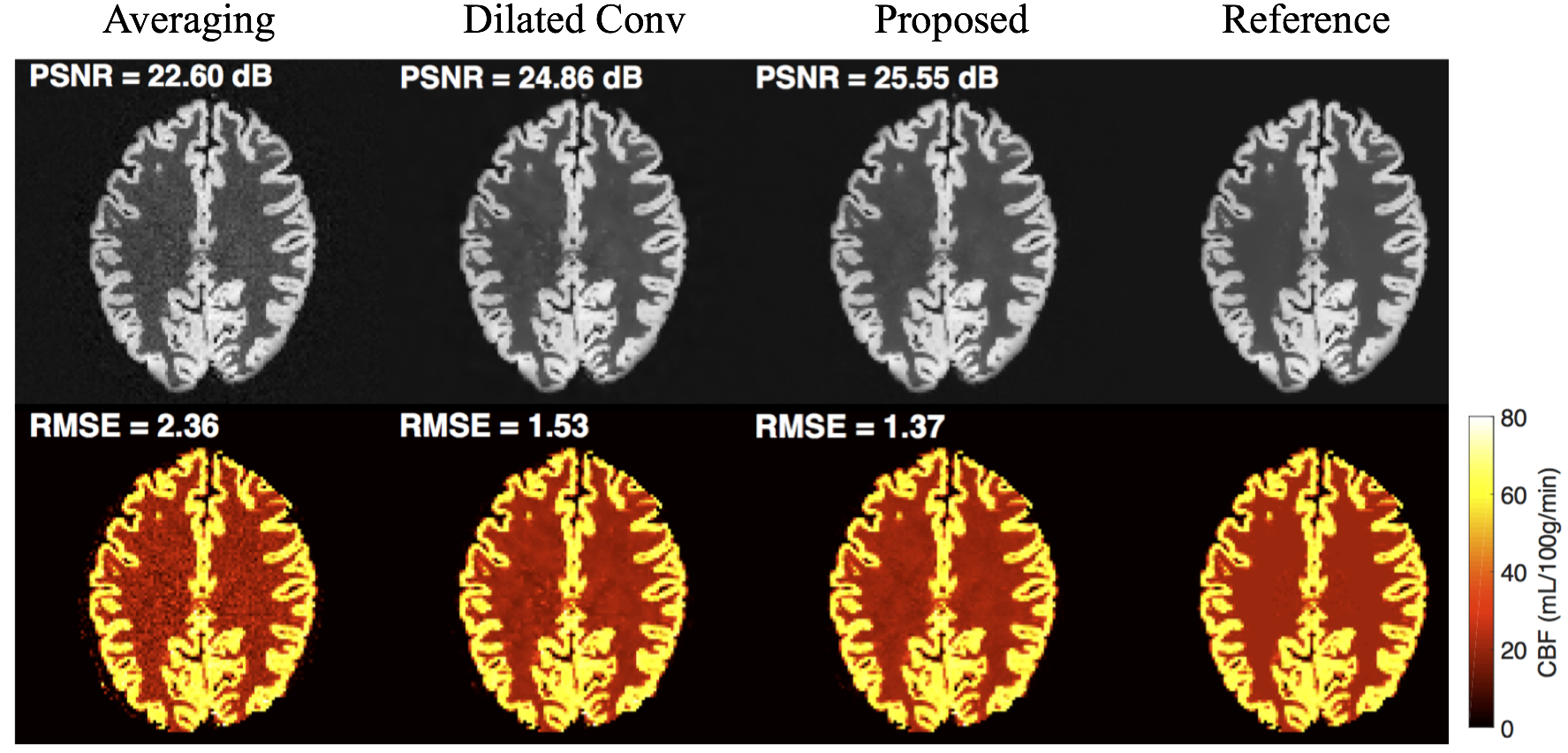}
 \caption{Visual comparison of denoising results (top) and resulting CBF maps (bottom) on an examplary synthetic data using $20\%$ of $100$ pairwise subtractions. Corresponding PSNR and RMSE values calculated with respect to references are also displayed at top-left corner of each image estimate. The proposed method can yield the best results both qualitatively and quantitatively.\label{fig:SyntheticResult}}
\end{figure}

% \vspace{-3mm}
\subsubsection{Data Preprocessing.}
\label{sec:preprocessing}
Prior to training the network, the standard preprocessing steps (motion correction, co-registration, Gaussian smoothing with $4$ $\text{mm}$ kernel size) \cite{Alsop} were applied on C/L pairs using our in-house toolbox implementation for ASL analysis. The top and bottom slices of each subject were removed from the analysis due to excessive noise caused by motion correction. 

Data augmentation was applied on every 2D image slices using rigid transformations. After augmentation, every image was divided into non-overlapping 2D patches of size $40 \times 40$, leading to $5440$ patches per subject. This process was repeated for input, target, and other variables required for network training. 

For each subject, we consider four different noise levels obtained by averaging randomly selected $20\%$, $40\%$, $60\%$ and $80\%$ of all available C/L repetitions, all of which were used during training and also tested on the trained network.

% \vspace{-4mm}
\subsubsection{Experimental Setup.}
\label{sec:experimentalsetup}
All experiments were performed using the leave-one-subject-out fashion. In order to show the benefit of our proposed method, we compare it with the recent deep learning based denoising method \cite{Kim} for ASL. Throughout the paper we refer to this method as \textit{Dilated Conv}. We evaluate this method using mean-squared-error (MSE) loss during training as proposed in the paper. We employ the peak signal-to-noise ratio (PSNR) to quantitatively assess the quality of image denoising, and the root-mean-squared error (RMSE) and Lin's concordance correlation coefficient (CCC) to assess the accuracy of CBF parameter estimation. We run the experiments on a NVIDIA GeForce Titan Xp GPU, and our code was implemented using Keras and TensorFlow libraries.

\begin{figure}[t!]
 \centering
 \includegraphics[width=\columnwidth]{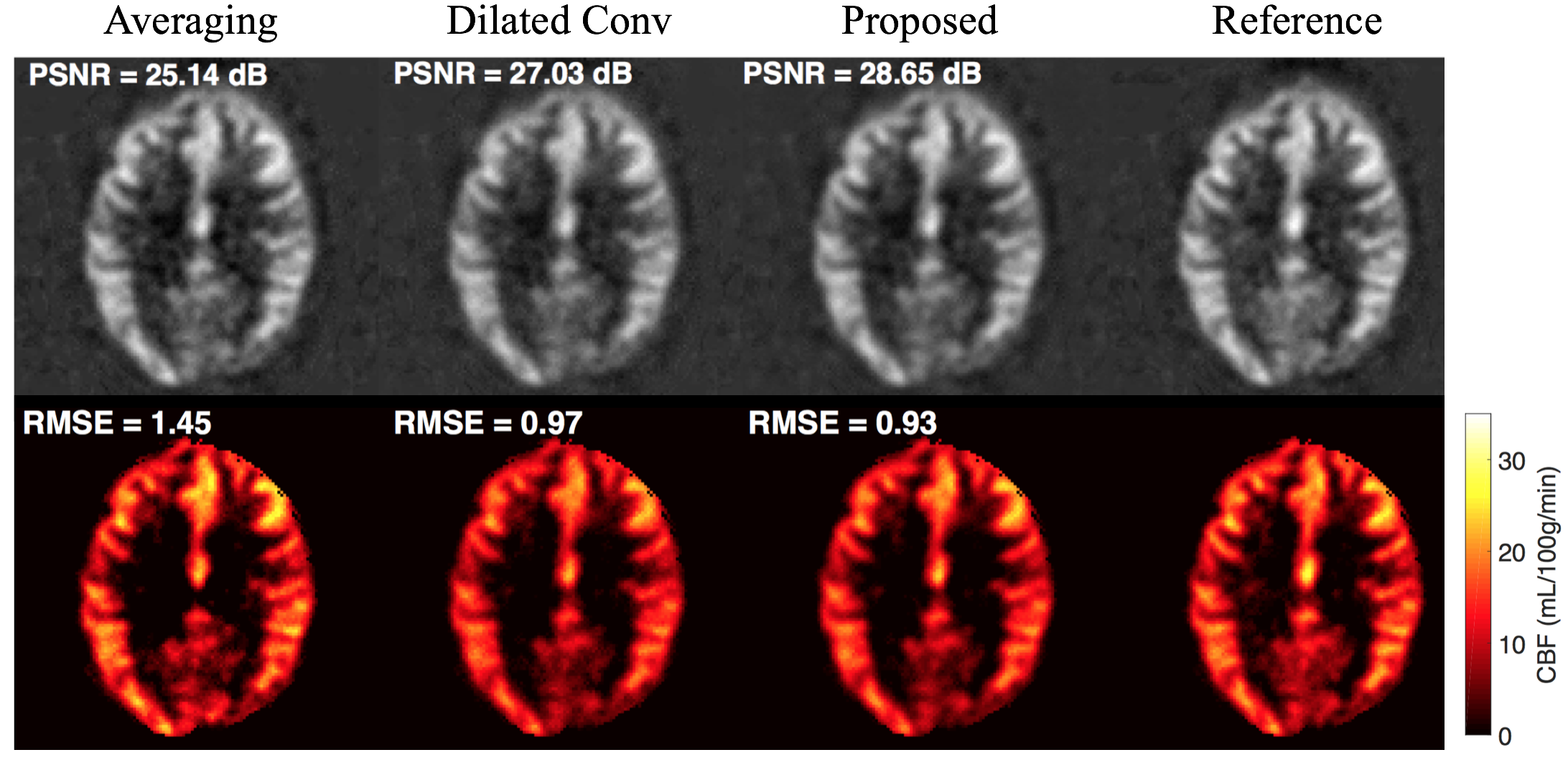}
 \caption{Visual comparison of denoising results (top) and resulting CBF maps (bottom) on an examplary real data using $40\%$ of $30$ pairwise subtractions. Although the estimated images qualitatively look similar, the quantitative metrics calculated inside the brain demonstrates the better performance of the proposed method.  \label{fig:RealResult}}
\end{figure}

% \vspace{-5mm}
\subsubsection{Results.}
\label{sec:results}
Figure~\ref{fig:SyntheticResult} demonstrates the denoised images and corresponding CBF maps of an exemplary slice of a synthetic dataset. Here, only $20 \%$ of $100$ synthetic C/L subtractions were used. Our proposed model produces the highest quality perfusion-weighted images where noise inside the brain is significantly removed compared to conventional averaging. The resulting CBF map of our proposed method is also closer to the reference CBF map yielding the lowest RMSE score.

\begin{figure}[t!]
 \centering
 \includegraphics[width=0.95\columnwidth]{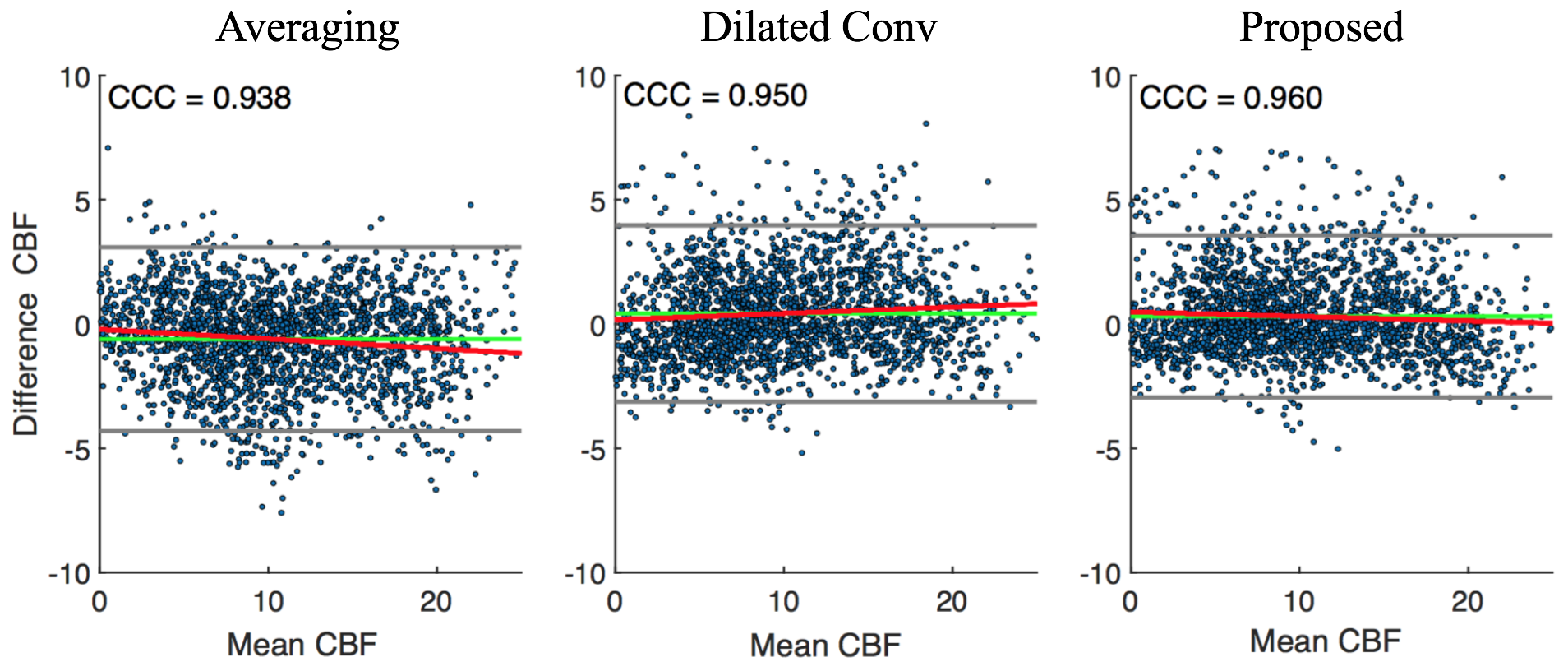}
 \caption{Bland-Altman plots of different methods obtained in a grey-matter region of a real subject's data. Differences in CBF (y-axis) between the reference and compared method is plotted against the mean (x-axis) values of the  two. The unit for horizontal and vertical axes are in $\mathrm{ml/100g/min}$. Solid green lines indicate mean difference. Solid gray lines at top and bottom correspond to upper and lower margins of $95\%$ limits of agreement. Linear regression lines are also shown with red solid lines. Corresponding CCC values are displayed at top-left corner of each plot.   \label{fig:BlandAltman}}
\end{figure}

\setlength{\tabcolsep}{0.43em}

\begin{table}[b!]
\centering
\caption{Quantitative evaluation in terms of $\mathit{mean(std)}$ obtained by different methods using all the subjects for synthetic and real datasets. The best performances are highlighted in bold font. All the metric values are calculated inside the brain region. Note that PR-$\lambda= x$ denotes our proposed method when $\lambda$ value is set to $x$.}
  \begin{tabular}{lcccccc}
    \toprule
    \multirow{2}{*}{\textbf{Method}} &
      \multicolumn{3}{c}{\textbf{Synthetic Dataset}} &
      \multicolumn{3}{c}{\textbf{Real Dataset}} \\
      \cmidrule(r){2-4}  \cmidrule(r){5-7} 
      & {PSNR} & {RMSE} & {CCC} & {PSNR} & {RMSE} & {CCC} \\
      \midrule
    Averaging & 20.3(4.81) & 2.20(2.51) & 0.88(0.06) & 23.6(6.16) & 1.49(0.80) & 0.85(0.07) \\
    Dilated Conv & 25.2(5.09) & 1.48(1.06) & 0.93(0.05) & 24.2(5.90) & 1.41(0.72) & 0.87(0.06) \\
    PR-$\lambda=0.2$ & \textBF{28.0(3.82)}& \textBF{1.33(0.79)} & \textBF{0.95(0.04)} & \textBF{25.1(5.36)}& \textBF{1.37(0.65)} & \textBF{0.88(0.05)} \\
    PR-$\lambda=0.5$ & 26.9(4.23) &1.40(0.84) & \textBF{0.95(0.04)} & 24.3(5.35) &1.38(0.67) & \textBF{0.88(0.06)}\\
    PR-$\lambda=0.7$ & 25.6(6.07) & 1.51(1.76) & 0.94(0.06) & 24.0(5.39) & 1.39(0.69) & \textBF{0.88(0.06)} \\
    PR-$\lambda=1.0$ & 25.3(5.62) & 1.49(1.56) &0.93(0.05) & 23.9(6.00) & 1.42(0.70) & 0.87(0.06) \\
    \bottomrule
  \end{tabular}
  \label{tab:quantTable}
\end{table}

In Fig.~\ref{fig:RealResult} we present the qualitative results from a real subject's data using $40\%$ of 30 C/L subtractions. Although the proposed method achieves the best PSNR and RMSE for perfusion-weighted image and CBF map respectively, the improvement against conventional averaging is less apparent compared to the synthetic data. The underlying reason is that as it can be clearly seen in Fig.~\ref{fig:RealResult}, our reference perfusion-weighted images obtained by averaging all 30 C/L subtractions still suffer from significant noise and artifacts. Since we train our network using these images as target, the network cannot produce results that show better quality beyond the reference images. The Dilated Conv method also faces similar problem for real data. 
Figure~\ref{fig:BlandAltman} depicts the Bland-Altman plots of CBF values in GM tissue obtained from different methods using a real subject's data. The plots indicate that our proposed method can yield better fidelity of CBF estimation with smaller bias (green solid line) and variance (difference between solid grey lines). The linear regression line (solid red) fitted in the averaging method also shows a systematic underestimation error whereas this error is considerably reduced by the proposed method where the regression line is closer to a straight line, $y=0$. Note that all three methods contain outlier voxels caused due to excessive noise and artifacts observable in most of the C/L subtractions.

We also quantitatively compare the predicted results in Table~\ref{tab:quantTable} in terms of PSNR, RMSE and CCC. Our proposed method outperforms other competing methods in all the metrics when either $\lambda=0.2$ or $\lambda=0.5$, which further demonstrates the advantage of the incorporation of CBF estimation model in denoising step. Taking into account data from all subjects, the differences between PR-$\lambda = 0.2$ and the Dilated Conv method on synthetic dataset are statistically significant with $p\ll0.05$ for all metrics. The differences are also statistically significant on real dataset for PSNR and RMSE, but not significant for CCC with $p=0.1388$.
Finally, we emphasize that image denoising using our trained network takes approximately $5$ ms on a single slice of matrix size $128\times 128$.

% \vspace{-2mm}
\section{Conclusion}
\label{sec:DiscussionAndConclusion}
We have proposed a novel deep learning based method for denoising ASL images. In particular, we utilize the Buxton kinetic model for CBF parameter estimation as a separate loss term where the agreement with reference CBF values is simultaneously enforced on the denoised perfusion-weighted images. Furthermore, we adopt the residual learning strategy on a deep FCN which is trained to learn a single model for denosing images from different noise levels. We have validated the efficacy of our method on synthetic and real pCASL datasets. %We anticipate that our model can yield improved performance on clinical ASL dataset containing high SNR and clean perfusion-weighted images to be used as a good reference. 
Future work will aim at extending our work to perform denoising on multi-TI ASL data where the estimation of the arterial transit time (ATT) parameter can be also exploited in the loss function.

\subsubsection{Acknowledgements.}
	%%This work was funded by  *** \\ ***
The research leading to these results has received funding from the European Unions H2020 Framework Programme (H2020-MSCA-ITN- 2014) under grant agreement no 642685 MacSeNet.

% Discussion points
%%%%%%%

%\subsubsection{Acknowledgements.}
	%%This work was funded by  *** \\ ***
%This work was funded by the European Commission under Grant Agreement Number 605162.

%
%
%%%%%%%%%%%%%%%%
% Bibliography
%%%%%%%%%%%%%%%%
%\vspace{-4mm}
\bibliographystyle{splncs03}
\bibliography{references}

\end{document}